\documentclass[conference,a4paper]{preprint}
\IEEEoverridecommandlockouts

\usepackage{soul}

\usepackage{amsmath,amssymb}
\usepackage{booktabs}
\usepackage{graphicx}
\usepackage{multirow}
\usepackage{xcolor}
\usepackage[hidelinks]{hyperref}
\usepackage{algorithm}
\usepackage{algpseudocode}

\usepackage{svg}
\svgsetup{inkscapelatex=false}

\usepackage[top=2cm, bottom=3cm, left=1.5cm, right=1.5cm]{geometry}

\newcommand{\Gsimple}{\ensuremath{G_{\mathrm{simple}}}}
\newcommand{\Gcurated}{\ensuremath{G_{\mathrm{curated}}}}
\newcommand{\Gcomplex}{\ensuremath{G_{\mathrm{complex}}}}

\title{StackGen3D: Stable 3D Block Stack Generation\\
from Multiview Silhouettes\thanks{A poster of this work is publicly available on the website of the
\href{https://sites.google.com/view/creative-assembly-workshop}{Creativity in Assembly Workshop at IROS 2026} (non-archival).}}

\usepackage{xspace}
\newcommand{\name}{\texttt{StackGen3D}\xspace}

\author{
\IEEEauthorblockN{
Xiaopeng Zhang\textsuperscript{1,*},
Luzhe Sun\textsuperscript{2,*},
Yiqing Xu\textsuperscript{3},
and Matthew R. Walter\textsuperscript{2}
}
\IEEEauthorblockA{
\textsuperscript{1}University of Chicago \quad
\textsuperscript{2}Toyota Technological Institute at Chicago \quad
\textsuperscript{3}Stanford University\\
xpengzhang0302@uchicago.edu,
luzhesun@ttic.edu,
yiqingx@stanford.edu,
mwalter@ttic.edu\\
\textsuperscript{*}Equal contribution.\\Project website: \url{https://ripl.github.io/StackGen3D-web/}
}
}

\begin{document}
\maketitle

% \begin{IEEEkeywords}
% Physical reasoning, Graph neural networks, Flow matching, Creative assembly.
% \end{IEEEkeywords}

\section{Introduction}

Creative assembly often begins from an incomplete specification rather than a fully defined 3D target: a user may provide a sketch, silhouette, or other abstract description of what the final structure should look like~\cite{sun2025stackgengeneratingstablestructures,xu2025setupfunctionalobject}. We study a concrete instance of this problem: given a fixed set of primitive blocks and three orthographic silhouettes, generate a physically stable 3D structure that matches the specified views.

This setting presents two coupled challenges. First, silhouette-to-3D generation is inherently underdetermined. Many block-pose configurations can produce nearly identical projections, while only a subset avoid interpenetration and remain stable. The goal is therefore to model a multimodal distribution over physically feasible structures consistent with the visual specification. Second, learning such a distribution requires a diverse dataset of stable 3D structures paired with silhouettes. Obtaining this data is difficult: physical feasibility must be evaluated for each candidate structure, while repeatedly searching and validating configurations with a physics simulator is expensive and difficult to scale. Simple procedural stacking heuristics provide stable examples, but cover only a limited portion of the space of possible structures.

We propose \name to address these challenges. First, we construct a scalable dataset of physically stable structures. We train a physics-informed graph neural network (PI-GNN) over block support relations as a fast surrogate for stability evaluation, and use it to curate diverse stable structures that are paired with multiview silhouettes. Second, using this dataset, we train a conditional flow-matching model to capture the multimodal distribution of block poses conditioned on the block inventory and target silhouettes. At inference time, lightweight AABB-based Best-of-$N$ selection or Remove-then-Rerun refinement reduces residual inter-block collisions without physics simulation.

Our preliminary results support both stages of the approach. First, our data-generation pipeline provides a strong basis for scalable curation of physically stable structures. A PI-GNN trained only on simple structures transfers substantially better to complex structures than a parameter-matched MLP (92.66\% vs.\ 46.86\% accuracy), indicating that support-relation reasoning generalizes across structural complexity and can serve as an effective stability model. Second, the generative model trained on the resulting stable-structure dataset learns to produce structures that satisfy the target silhouettes while capturing useful physical regularities. A single flow-matching sample achieves 0.967 silhouette IoU, and simulator-free Best-of-10 selection increases post-hoc physical stability from 45.3\% to 72.4\% while preserving silhouette fidelity. Together, these results provide preliminary evidence that scalable stable-data generation can enable learned generative models for underdetermined creative structure.

\begin{figure}[!t]
    \centering
    \includegraphics[width=0.70\columnwidth,trim={0 11 0 11},clip]{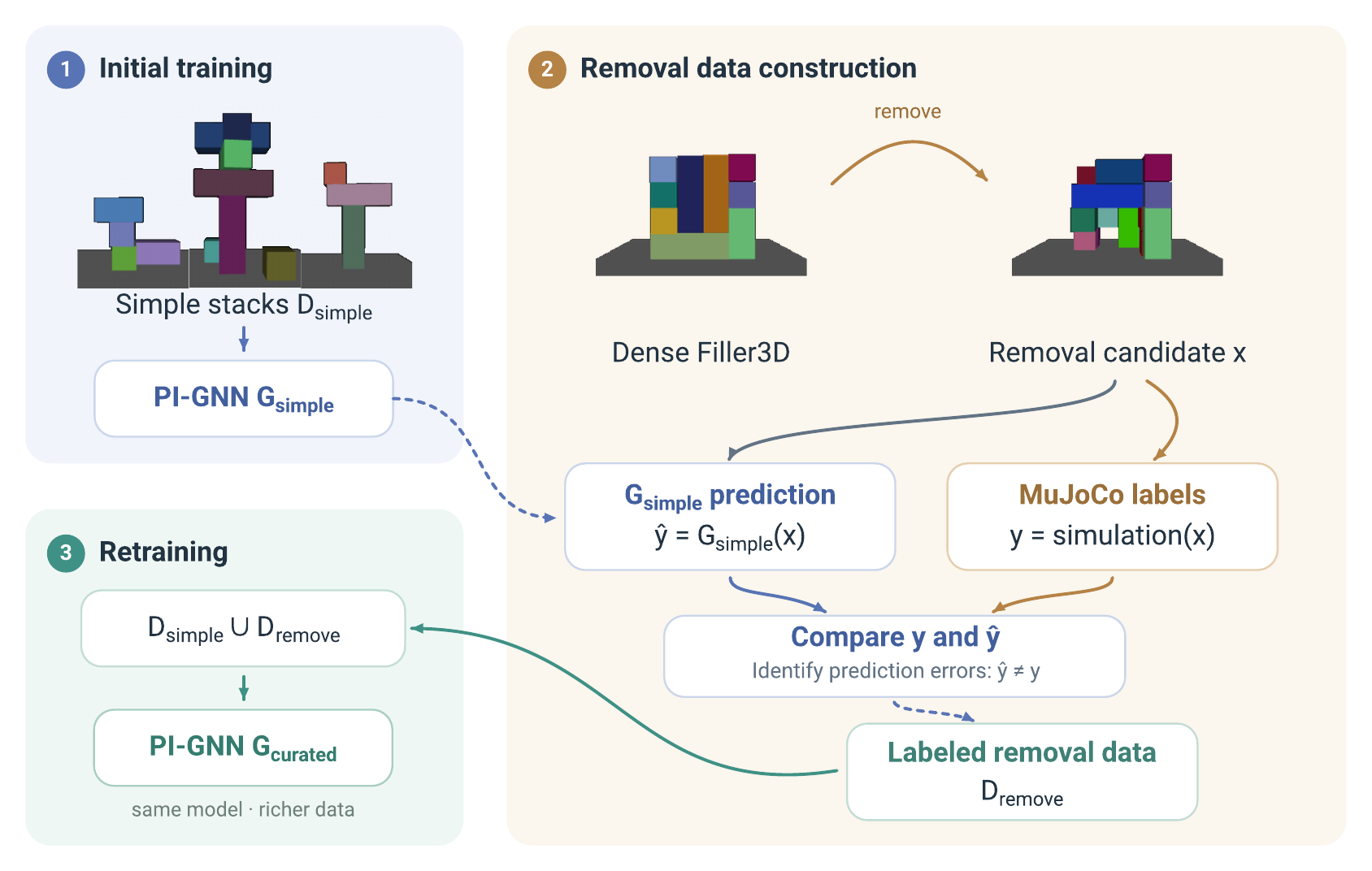}
    \vspace{-2mm}
    \caption{PI-GNN Training Process.}
    \vspace{-4mm}
    \label{fig:gnn_pipeline}
\end{figure}

\section{StackGen3D}

\name follows a two-stage structure: we first construct a large-scale dataset of physically stable structures, and then learn a conditional generative model from this dataset.

\subsection{Relational Stability and Dataset Curation}

Physical stability is inherently relational: whether a block remains stable depends on how it is supported by neighboring blocks and on the load transmitted through those supports. We therefore represent each structure as a directed support graph and use a PI-GNN~\cite{gilmer2017neural,fey2019pyg} to predict block-level stability and displacement. Nodes encode block geometry, pose, mass, and accumulated supported mass, while edges encode geometric support relations such as contact area and relative displacement. Message passing allows these local support interactions to propagate through the structure.

The larger challenge is obtaining diverse stable structures for generative-model training. As shown in Fig.~\ref{fig:gnn_pipeline}, we first train an initial PI-GNN $\Gsimple$ on balanced simple stacks generated by a COM-based heuristic. We then construct more complex candidates by randomly removing blocks from dense Filler3D structures and use MuJoCo~\cite{todorov2012mujoco} to obtain stability supervision. Combining these examples with the simple-stack data yields a predictor $\Gcurated$, which efficiently filters candidate structures during dataset construction. The stable structures are paired with multiview silhouettes, forming the training data.

\subsection{Conditional Flow Matching and Geometric Inference}

We next ask whether a generative model can learn the distribution of stable 3D structure compatible with a silhouette specification. Because the mapping is inherently one-to-many, we use conditional flow matching rather than predicting a single deterministic pose configuration.
Let $\mathbf x_1$ denote the block poses in a simulator-verified stable
structure, with each pose represented by a position and quaternion. Given
$\mathbf x_0\sim\mathcal N(\mathbf 0,\mathbf I)$ and
$t\sim\mathcal U[0,1]$, conditional flow matching uses
$\mathbf x_t=(1-t)\mathbf x_0+t\mathbf x_1,
\mathcal L_{\mathrm{FM}}
=
\mathbb E\!\left[
\left\|
v_\theta(\mathbf x_t,t;\mathcal S,\mathcal I)
-(\mathbf x_1-\mathbf x_0)
\right\|_2^2
\right]$.

At inference, we integrate the learned Transformer-based velocity field~\cite{vaswani2017attention}  from $t=0 \to 1$
with an ODE solver~\cite{chen2018neural}, conditioning on the block inventory
$\mathcal S$ and silhouettes $\mathcal I$. Generated structures may nevertheless contain inter-block penetration. 
Best-of-$N$ draws $N$ samples and selects the one with the smallest
maximum pairwise AABB penetration. 
Remove-then-Rerun uses the largest AABB overlap to select one block for removal,
updates $\mathcal S$, and resamples poses for the reduced block set until no
overlap exceeds a threshold or removing all blocks. Neither
procedure guarantees physical stability; physics simulation is used only afterward for
evaluation.

\section{Experiments}

\subsection{PI-GNN Stability Prediction}

\begin{table}[t] \centering 
\caption{Best-of-N generation (mean $\pm$ std. over 10 folds).} \label{tab:best of n} \scriptsize \setlength{\tabcolsep}{3.2pt} \begin{tabular}{lccc} \toprule Inference method & Stability (\%) & IoU & Penetration \\ \midrule Single sample ($N=1$) & $45.30\!\pm\!4.10$ & $.967\!\pm\!.002$ & $.711\!\pm\!.066$ \\ Best-of-5 & $66.70\!\pm\!4.61$ & $\mathbf{.970}\!\pm\!.002$ & $.134\!\pm\!.026$ \\ Best-of-10 & $\mathbf{72.40}\!\pm\!4.63$ & $\mathbf{.970}\!\pm\!.002$ & $.064\!\pm\!.016$ \\ Best-of-25 & $67.80\!\pm\!5.15$ & $\mathbf{.970}\!\pm\!.002$ & $.034\!\pm\!.005$ \\ Best-of-50 & $69.30\!\pm\!4.94$ & $.969\!\pm\!.002$ & $.028\!\pm\!.003$ \\ Best-of-100 & $66.00\!\pm\!4.05$ & $.969\!\pm\!.003$ & $\mathbf{.025}\!\pm\!.002$ \\ Remove-then-Rerun & $64.60\!\pm\!4.13$ & $.953\!\pm\!.004$ & $.099\!\pm\!.009$ \\ \bottomrule \end{tabular} 
\vspace{-5mm}
\end{table}

\begin{figure}[!t]
\centering
\includegraphics[width=.80\columnwidth,trim={0 0 0 36},clip]{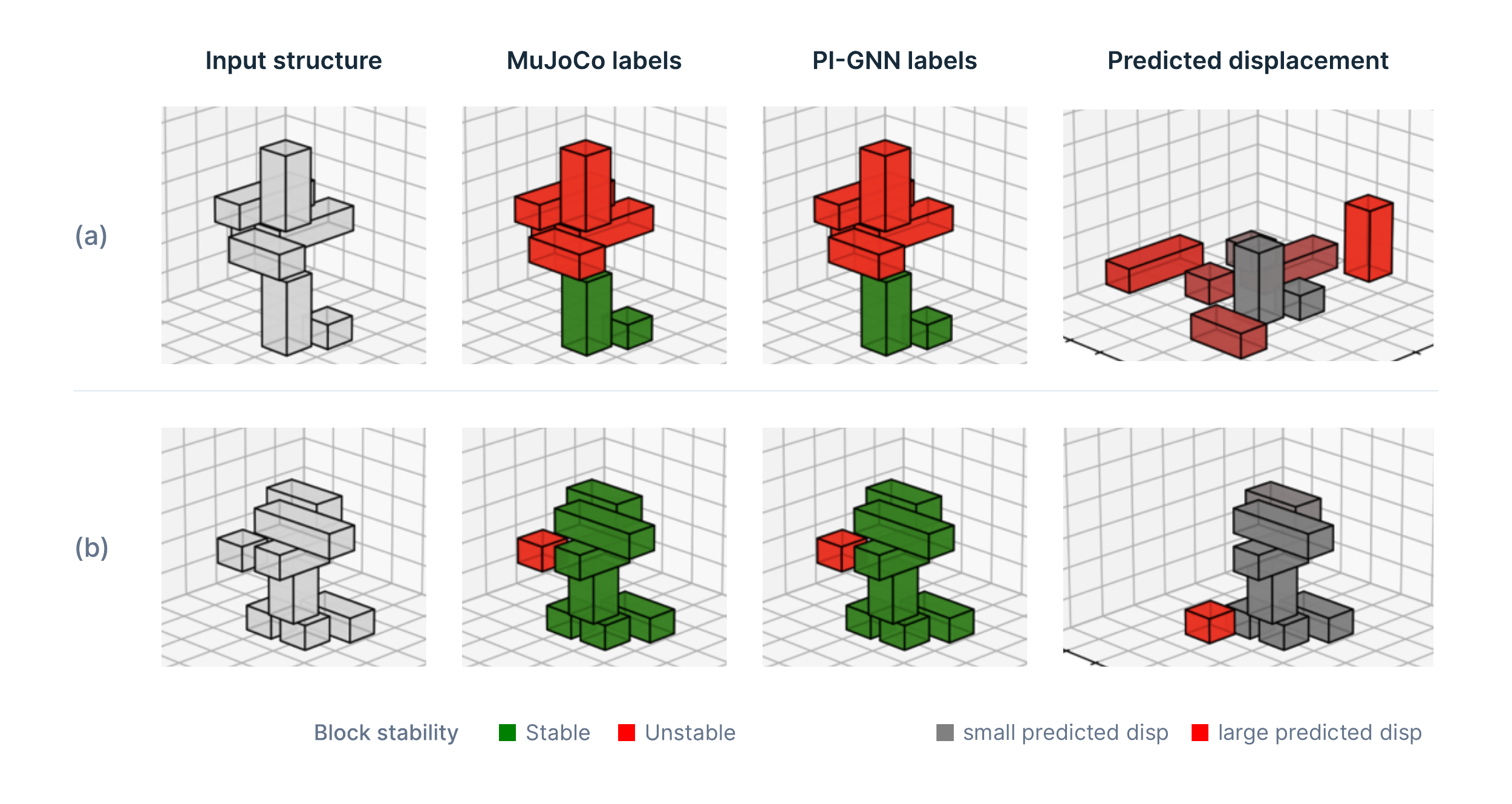}
\vspace{-7mm}
\caption{Qualitative PI-GNN stability predictions on complex structures. Block-level predictions are compared with MuJoCo labels, illustrating how learned support reasoning transfers from simple structures to complex.}
\vspace{-2mm}
\label{fig}
\end{figure}

We first evaluate whether relational stability reasoning can support our data-generation pipeline. We compare \Gsimple{} with a parameter-matched $MLP_\mathrm{simple}$ trained on the same simple structures, and with \Gcomplex{}, which uses the same PI-GNN architecture but is trained directly on complex structures. The PI-GNN and MLP contain 261,576 and 261,383 parameters, respectively. Because 85.1\% of blocks in the complex data are stable, we additionally report balanced accuracy,
$\mathrm{BAcc}=\tfrac{1}{2}(\mathrm{TPR}+\mathrm{TNR})$. Despite similar performance on simple structures, \Gsimple{} transfers substantially better to complex structures than the parameter-matched MLP (92.66\% vs.\ 46.86\% accuracy), supporting the importance of explicitly modeling support relations. On the same complex test set, \Gsimple{} approaches \Gcomplex{} (92.66\% vs.\ 97.58\%), although the balanced-accuracy gap shows that direct complex-domain training remains beneficial for identifying unstable blocks. Fig.~\ref{fig} provides qualitative examples of this transfer, showing that the PI-GNN can identify stable and unstable blocks in structurally complex structures. For reference, \Gcurated, trained on the combination of simple and
Filler3D/remove structures, achieves 96.32\% accuracy, 92.87\% balanced
accuracy, and 93.90\% macro-F1 on its combination validation set. We report
this as a reference rather than as part of the controlled
baselines.

\subsection{Structure Generation without Online Simulation}

\begin{figure}[!t]
\centering
\includegraphics[width=.80\columnwidth,trim={0 0 0 25},clip]{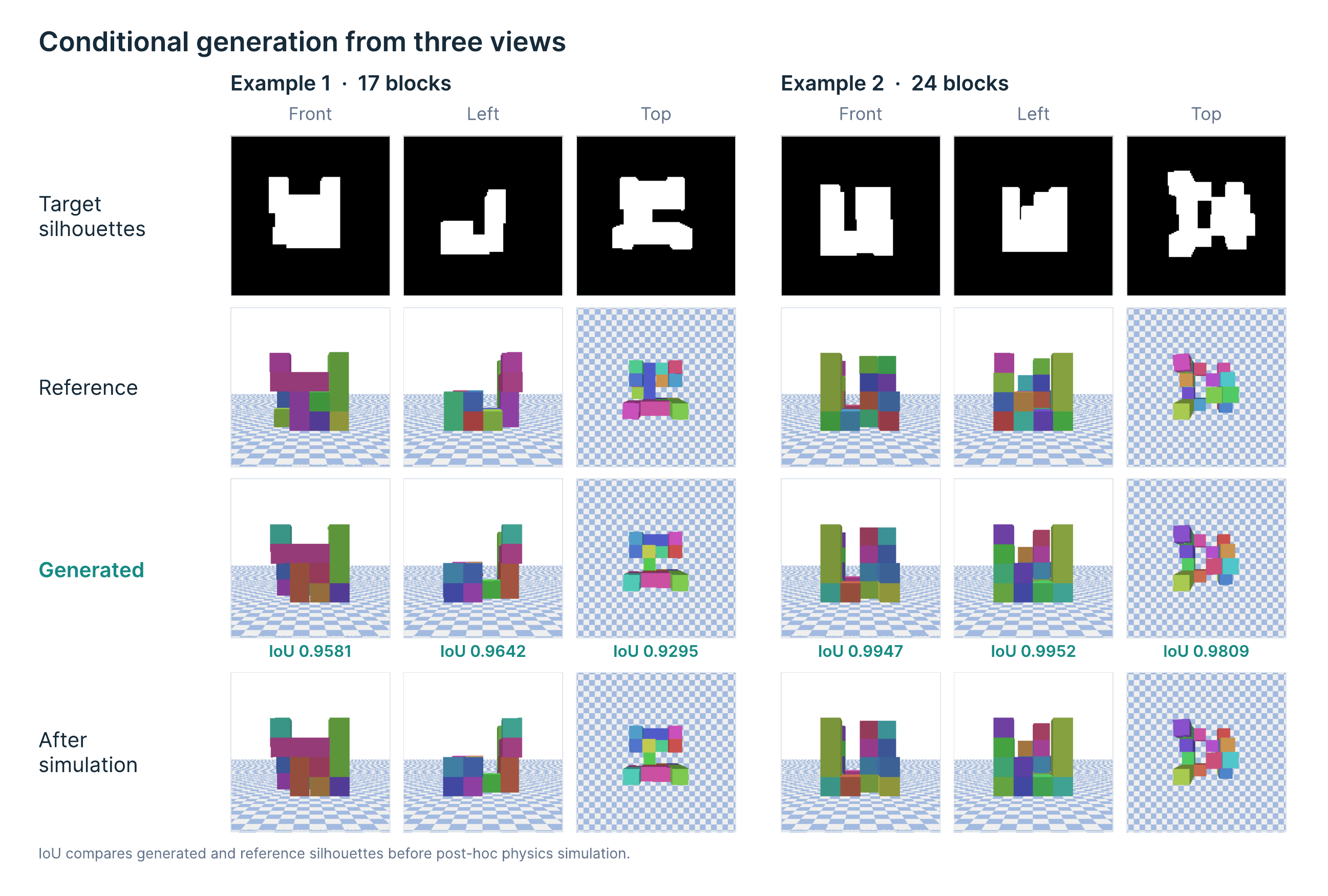}
\vspace{-5mm}
\caption{Qualitative conditional flow-matching generations. Each example shows the input orthographic silhouettes, reference structure, generated structure before simulation, and its configuration after post-hoc physics simulation.}
\vspace{-5mm}
\label{fig:generation}
\end{figure}

\begin{table}[!t] \centering \caption{Block-level stability-prediction baselines.} \label{tab:controlled} \scriptsize \setlength{\tabcolsep}{4.0pt} \begin{tabular}{@{}cllrrr@{}} \toprule Exp. & Training domain / model & Test domain & Acc. (\%) & BAcc. (\%) & F1 (\%) \\ \midrule \multirow{4}{*}{A} & Simple / MLP & Simple & 89.81 & 89.60 & 89.16 \\ & Simple / \Gsimple & Simple & 90.01 & 87.81 & 88.90 \\ & Simple / MLP & Complex & 46.86 & 67.22 & 45.05 \\ & Simple / \Gsimple & Complex & \textbf{92.66} & \textbf{86.12} & \textbf{85.65} \\ \midrule \multirow{2}{*}{B} & Simple / \Gsimple & Complex & 92.66 & 86.12 & 85.65 \\ & Complex / \Gcomplex & Complex & \textbf{97.58} & \textbf{93.76} & \textbf{95.07} \\ 
\midrule
Ref. & Combination / \Gcurated & Combination val.
& 96.32 & 92.87 & 93.90 \\
\bottomrule \end{tabular} 
\vspace{-5mm}\end{table}

We next evaluate whether the conditional flow model trained on the curated stable-structure dataset can generate structures that satisfy the target silhouettes while retaining useful physical regularities. We evaluate 10 folds with 100 samples per fold. Inference uses only AABB overlap, while PyBullet~\cite{coumans2021pybullet} is used afterward to evaluate physical stability. 
As shown qualitatively in Fig.~\ref{fig:generation}, the generated structures closely follow the input silhouettes, while post-hoc simulation reveals the remaining physical failure modes. Quantitatively, a single sample already achieves 0.967 silhouette IoU (Tab.~\ref{tab:best of n}), indicating that the learned conditional distribution strongly captures the geometric specification. Best-of-10 improves stability by $27.1\%$ over a single sample while leaving silhouette IoU essentially unchanged, showing that the learned distribution contains substantially more physically viable modes than are exposed by one sample. Remove-then-Rerun similarly improves stability with only a modest reduction in IoU. Beyond $N=10$, penetration continues to decrease while stability does not improve monotonically, indicating that AABB overlap is a useful but incomplete proxy for physical stability.

\bibliographystyle{IEEEtran}
\bibliography{references}

\clearpage
\appendices

\end{document}